# Careful analysis of XRD patterns with Attention

Koichi Kano[1], Takashi Segi,[1] and Hiroshi Ozono[2]

[1] *Computational Science Department, KOBELCO Research Institute, INC., Takatsuka-dai 1-5-5, Nishi-ku, Kobe, Hyogo 651-2271, Japan*

[2] *Electric Vehicle Evaluation Technology Section. KOBELCO Research Institute, INC., Takatsuka-dai 1-5-5, Nishi-ku, Kobe, Hyogo 651-2271, Japan.*

The important peaks related to the physical properties of a lithium-ion rechargeable battery were extracted from the measured X-ray diffraction spectrum by a convolutional neural network based on the Attention mechanism. Among the deep features, the lattice constant of the cathodic active material was selected as a cell-voltage predictor, and the crystallographic behavior of the active anodic and cathodic materials revealed the rate property during the charge–discharge states. The machine learning automatically selected the significant peaks from the experimental spectrum. Applying the Attention mechanism with appropriate objective variables in multi-task trained models, one can selectively visualize the correlations between interesting physical properties. As the deep features are automatically defined, this approach can adapt to the conditions of various physical experiments.

## 1.    Introduction

Material scientists usually extract the important peaks from measured spectral data, and relate them to the physical properties of an interesting substance. To investigate the cycle properties of $Li(Ni_x,Mn_y,Co_z)O_2$ (NMC) cathodic materials in lithium-ion secondary batteries, researchers have determined the amount of cation mixing from the (003)/(004) peak intensity ratio in the X-ray diffraction (XRD) pattern of the layered rock-salt type unit cell [1-2]. Similarly, the mechanical properties of amorphous carbon materials have been derived from the intensity ratio of the G and D bands (at approximately 1560 and 1350 cm$^{-1}$, respectively) in the Raman spectra of the materials [3-5]. Significant peaks are selected by carefully studying the causal relationship between the properties and spectra of functional materials. The spectra can be derived from many physical phenomena, such as X-rays, neutron rays, electron rays, gamma rays, nuclear magnetic resonance, and mass spectrometry. Therefore, when extracting peaks with causal relationships to physical properties, the researcher requires deep expertise in both materials science and physics.

Today, new functional materials are often found by inverse-problem solving, screening, and surveying with machine learning techniques [6-8]. Such techniques also improve the quality of measurement data [9-11]). The important peaks in the spectra are then probed by data-driven approaches. In one approach, the materials scientist defines the amount of a descriptor expressed in a specific spectrum. Typical spectral descriptors are peak positions, half widths, and peak intensities. Feature selection by regression analysis then identifies the primary ingredients that correlate with interesting physical properties. This approach is familiar to materials scientists, but requires the preparation of dedicated descriptors for numerous measurement technologies. For example, the peak position in an XRD pattern is essential for determining the crystallographic data of the compounds in a sample. Meanwhile, identification by fingerprint collation, like that the X-ray absorption and electronic loss near the edge structure, requires not the peak information but the peak shape of the spectrum. Namely, descriptors must be expressed in various formats. Another approach uses deep learning techniques such as the convolutional neural network (CNN) developed by Neocognitron [12], which is based on the recognition mechanism in the visual cortex of the brain [13]. Besides image recognition, CNNs can analyze one-dimensional data such as spectra [14]. The main advantage of deep learning is that the above-mentioned descriptors are automatically determined. However, the deep features are generally difficult to interpret because they are expressed in a nested nonlinear structure. Many data scientists have attempted to create readable deep features through artificial intelligence [15]. For example, class average maps (CAMs) are widely used in CNNs and similar learning-based methods. Oviedo *et al.* recently clarified the cause of misclassification in the XRD patterns of 115 metal–halogen compounds [16]. A method called *Attention* has greatly improved the learning accuracy of natural language processing [17]. Lin *et al.* showed that *Attention* visualizes the words that are important in context. And we emphasize that *Attention* not only improves the accuracy of natural language processing [18-20], but also enables visualization of the prediction basis [21]. In this way, *Attention* can visualize the important information by considering the correlations in the data. Attention considers the data as dictionary objects [22] and obtains their correlations using inner products [23]. Combinations of CNN and *Attention* are commonly reported in the literature. An example is the super-resolution problem [24-26], which learns the spatial correlations between channels and the correlations between global and spatial information. After learning the global–spatial correlations, the region of interest, called the *Attention* mask, can be visualized [27].

In the present work, the important peaks in the experimental spectra of a lithium-ion secondary battery were automatically extracted by a data-driven approach. First, the diffraction pattern and battery voltage were collected by in-situ XRD measurements. Next, a CNN model was trained on approximately 4,000 experimental results. Finally, the deep features were visualized and projected onto the diffraction patterns, and correlated with the cell properties of the cathode, anode and current collector foils.

**2. Method**

We first describe the experimental method in detail. The purpose of this experiment was to acquire the information of the XRD pattern as the explanatory variable and the voltage as the objective variable. All measurements were performed in-situ on a lithium-ion secondary battery. This lithium-ion battery (LIB) was a pouch-type cell composed of an NMC positive electrode and a graphite negative electrode. The fabrication of the sample is described elsewhere [28]. The thickness was designed to enable transmission-mode XRD with Cu-$K\alpha$ radiation. The XRD equipment was a SmartLab diffractometer with a Pilatus 2D-detector (Rigaku Corporation, Tokyo, Japan). In the absence of mechanical operation by a goniometer or scintillation counter, the acquisition time of the diffraction pattern was several seconds. The measurement angle (Bragg angle 2θ) was ranged from 20 to 40°. During the XRD measurements, the LIB voltage was controlled between 2.6 and 4.2 V by a charge–discharge test module at room temperature. The charge–discharge rates were 1.0 and 0.2 C, where 1.0 C means that the state transits from fully discharged to fully charged in one hour. The XRD pattern and charge–discharge curves at 1.0 and 0.2 C are depicted in Fig. 1.

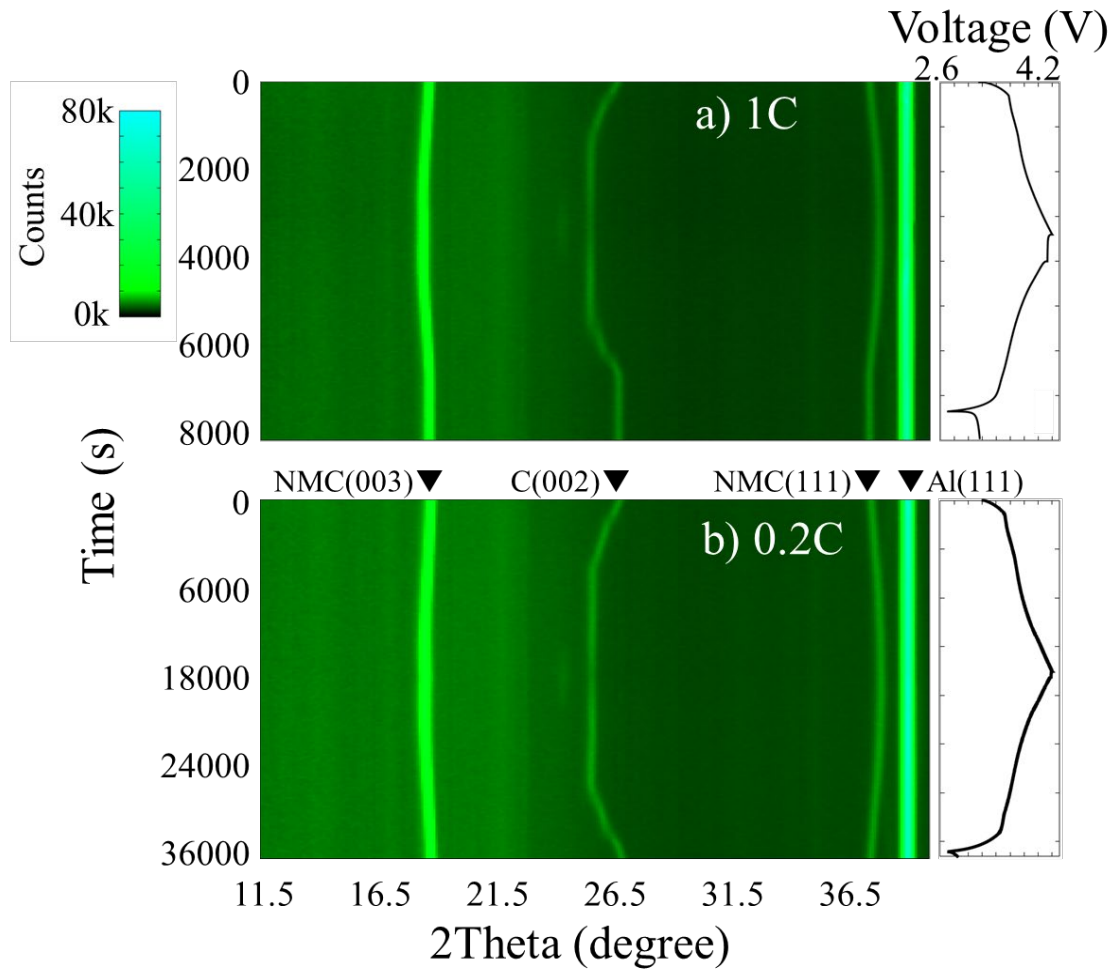

**Fig. 1.** XRD pattern and charge–discharge curves at 1 C and 0.2 C,
obtained from in-situ measurements of the lithium-ion secondary battery.

The LIB returned to the discharged state after reaching the fully charged state. The diffraction patterns contained four important peaks. The NMC (003) and (111) peaks were derived from the positive electrode, the C (002) peak was attributable to graphite in the negative electrode, and the Al (111) peak was contributed by aluminum. The NMC peaks shifted because delithiation changed the lattice constant of the NMC. The C (002) peak of the negative electrode is important because lithium ions are stored in the van der Waals gap of graphite, thus changing the lattice constant of the anode materials. Conversely, the aluminum (111) peak was independent of the charge/discharge state because aluminum comprises only the current collector foil of the positive electrode, so peak broadening and peak shifts were not expected.

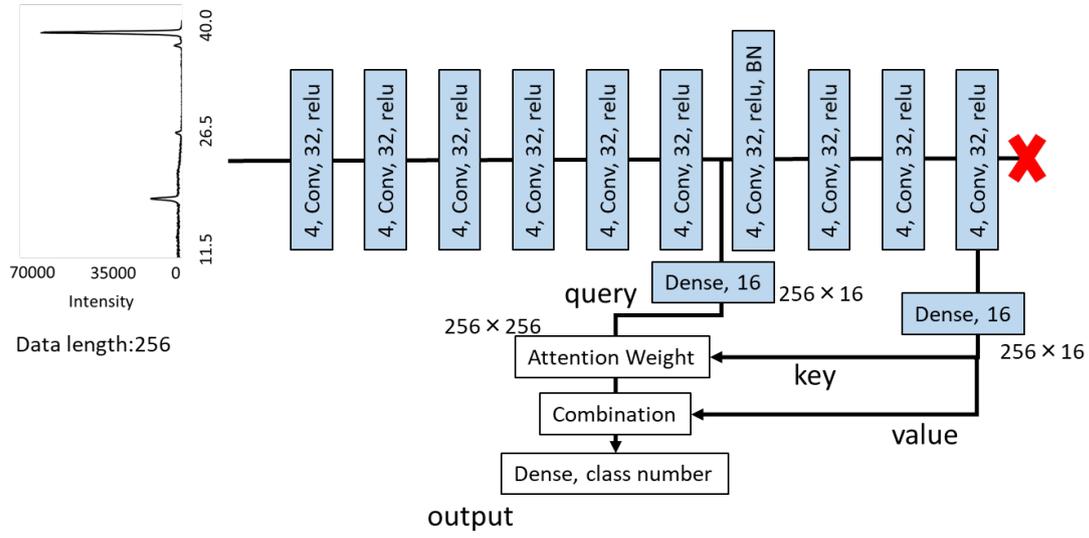

**Fig. 2.** Flowchart of the CNN and Attention processing of the experimental data.

Figure 2 shows the automatic extraction process of the important peaks related to the physical properties of the LIB. The first step trains the CNN model for predicting the objective variable (voltage) from the explanatory variable (XRD pattern). The peak intensities were standardized by the maximum and minimum photon counts in the Bragg diffraction. The 1400 experimentally obtained vectors were resized to 256-dimensional vectors by spline interpolation. The vector data were processed through 10 layers of the CNN. The CNN was operated with 32 filters, four kernel sizes, and a rectified linear unit activation function. Batch normalization was applied to the seventh layer of the CNN. Next, the deep features were visualized with single-head Attention. The output value of the sixth CNN layer was defined as the QUERY and the tenth output value was defined as the KEY and VALUE. The Attention Weight was then calculated as

$$U_{Attention\ Weight} = f(\mathbf{Q} \otimes \mathbf{K}^T), \quad (1)$$

where $\mathbf{Q}$ and $\mathbf{K}$ are the tensors obtained from the above-mentioned outputs of the CNN layers. The dot product of $\mathbf{Q}$ and $\mathbf{K}$ was computed by a softmax activation function. Combination (see Fig. 2) was defined as the dot product of VALUE and Attention Weight using a hyperbolic tangent activation function.

This model predicted the LIB voltage from the XRD pattern. However, the relationship between the actual charge–discharge states and the diffraction patterns was influenced by hysteresis. For example, the diffraction pattern obtained at 4 V during the charging process differed from that of the same voltage measured during the discharging process. This problem was resolved by defining three target variables: *voltage*, *mode*, and *rate*. The *voltage* corresponded to the operating potential of the LIBs, and *mode* referred to the charging state (first or second half of the charging or discharging process). The *rate* represented the normal (1.0 C) or slow (0.2 C) charge–discharge rate. Note that *mode* and *rate* were categorical data.

The deep features of the relation between the XRD pattern and the LIB properties were visualized by three prediction models. In Case 1, the LIB voltage was predicted from the experimental results at 1.0 C. Case 2 was similar to Case 1, but the *voltage* and *mode* were predicted simultaneously (i.e., predicted in a multi-tasking situation). In Case 3, the *voltage*, *mode*, and *rate* were predicted simultaneously from the experimental results at 1.0 C or 0.2 C. In Cases 2 and 3, the multiple output layers were prepared after the Combination block shown in Fig. 2. As the loss functions, we applied an inverse hyperbolic cosine function for voltage, and cross-entropy functions for the mode and rate predictions. The error during training was expressed as a linear combination of the three weighted loss functions. The relationship between the voltage and the XRD pattern was investigated from 3.6 to 4.6 V. The 2.6–3.6 V region was excluded because it was dominated by electrolyte diffusion rather than by crystallographic alteration of the NMC and graphite materials. Approximately 4,000 XRD patterns and redox potentials of the charge–discharge LIB tests were collected at each charging rate (1.0 and 0.2 C). Half of these data were employed as the training data; the remainder were preserved for verifying the deep learning prediction by CNN.

3. **Results and discussion**

Figure 3 compares the predicted (red) and experimental (black) voltages as function of time in each case. Table I gives the average absolute error (MAE) in the predicted *voltage*, and the categorical accuracies of *mode* and *rate*. In all cases, the trained models predicted the XRD pattern and charge–discharge characteristics with sufficiently high accuracy.

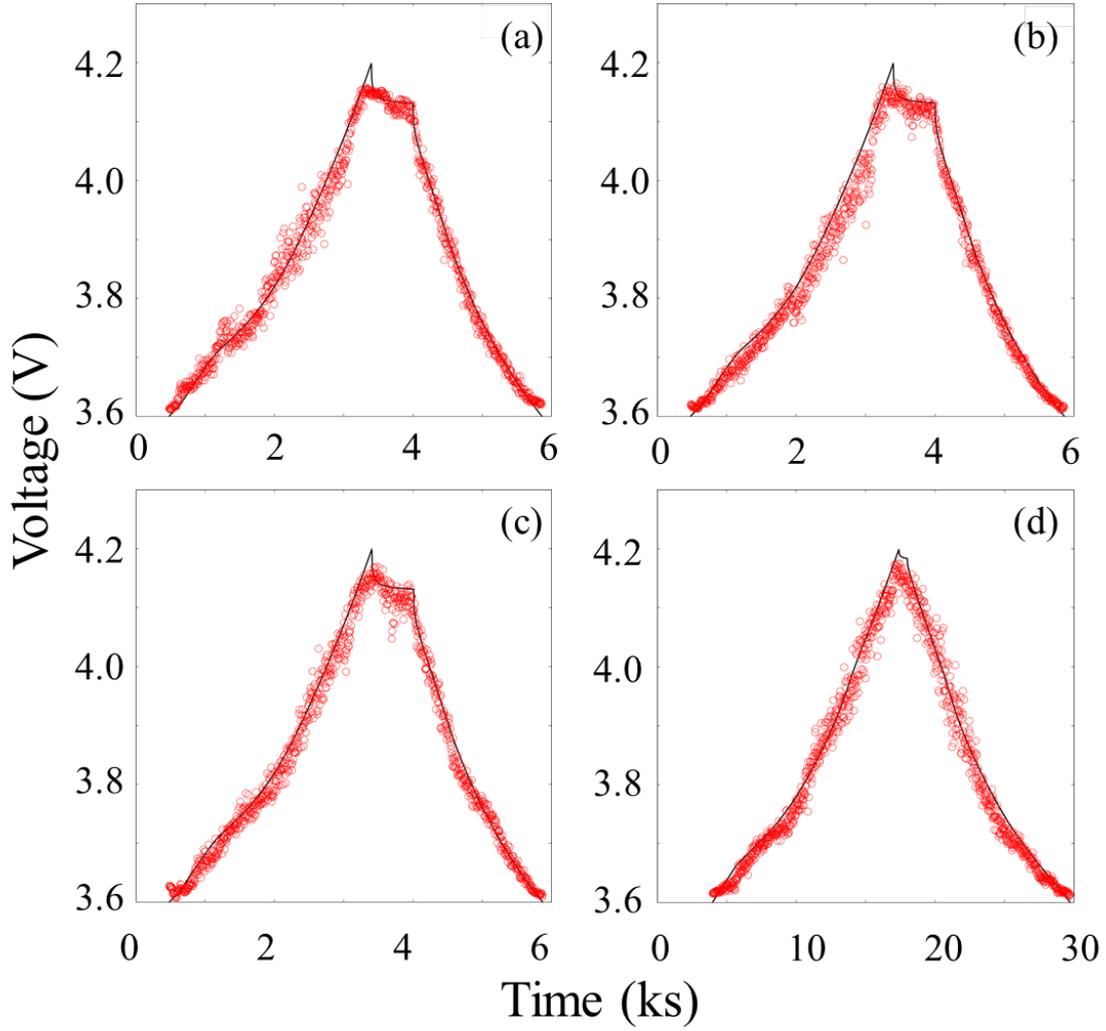

**Fig. 3.** Comparison of predicted and experimental results (red open circles and solid lines, respectively). The predicted values were obtained from our CNN models. Shown are the results of (a) Case 1 (single task prediction at 1.0 C), (b) Case 2 (two-task prediction at 1.0 C), (c) Case 3 (three-task prediction) at 1.0 C, and (d) Case 3 (three-task prediction) at 0.2 C.

**Table I.** Prediction accuracy in each case

| Case | *Voltage* (MAE) | *Mode* (%) | *Rate* (%) |
|---|---|---|---|
| Case 1 | 1.47E–2 | - | - |
| Case 2 | 1.86E–2 | 97 | - |
| Case 3 | 1.89E–2 | 98 | 100 |

The deep features of the trained model were visualized by *Attention*. To reveal the relationship between the XRD peaks and LIB properties, the classification and regression problems were solved together, which forbids a straightforward adaptation of the CAM scheme. In the present study, the applications of *Attention* were expected to be extendable from natural language processing to applied physics problems. The *Attention Weight* was expressed as a two-dimensional (256 × 256) matrix with axes of QUERY and KEY. As the XRD pattern was a 256-dimensional vector, the maximum KEY value of each QUERY was obtained. The one-dimensional *Attention Weight* was defined as the *Visualized Attention Weight* (VAW). Figures 4 and 5 show the XRD patterns and VAW projection results at 1.0 C (all cases) and 0.2 C (Case 3), respectively, with the VAW values projected onto the XRD pattern. The *Attention Weight* values were standardized from 1.0 (completely black) at maximum to 0.0 (completely white) at minimum. The black regions are the regions of interest in the predictions of each case.

In Cases 1 and 2, the VAW was high near the NMC (003) peak, but the graphite (002) peak can be ignored. Self-evidently, the voltage prediction needs the crystallographic data of the positive electrode, but not that of the graphite materials. Conversely, when predicting both the *voltage* and *rate* (Case 3 of Figs. 4 and 5), the VAW was high at the peaks of both the cathodic and anodic materials. As the number of Li ions increased in the van der Waals gap, structures such as $LiC_6$ and $LiC_{12}$ appeared in the negative graphite electrode. According to previous studies, Li-ion imbalance in the matrix is enhanced at high charge and discharge rates, causing peak broadening in the diffraction patterns [29]. Therefore, it is important that the behavior of both active materials.

Unfortunately, *Attention* also focused on the aluminum peaks in all cases, and on the 11.5° angle of the VAW map (Figs. 4 and 5). When preparing the trained model, the information of the *d*-spacing and the X-ray wavelength were not provided. Therefore, the deep features of our CNN models relatively judged the positions of the most prominent peaks; namely, the Al (111) peak and the both ends of the diffraction pattern.

The visualized deep-features correlation between the explanatory variable and the objective variable contains the important factors that correspond to the domain knowledge of lithium-ion secondary batteries.

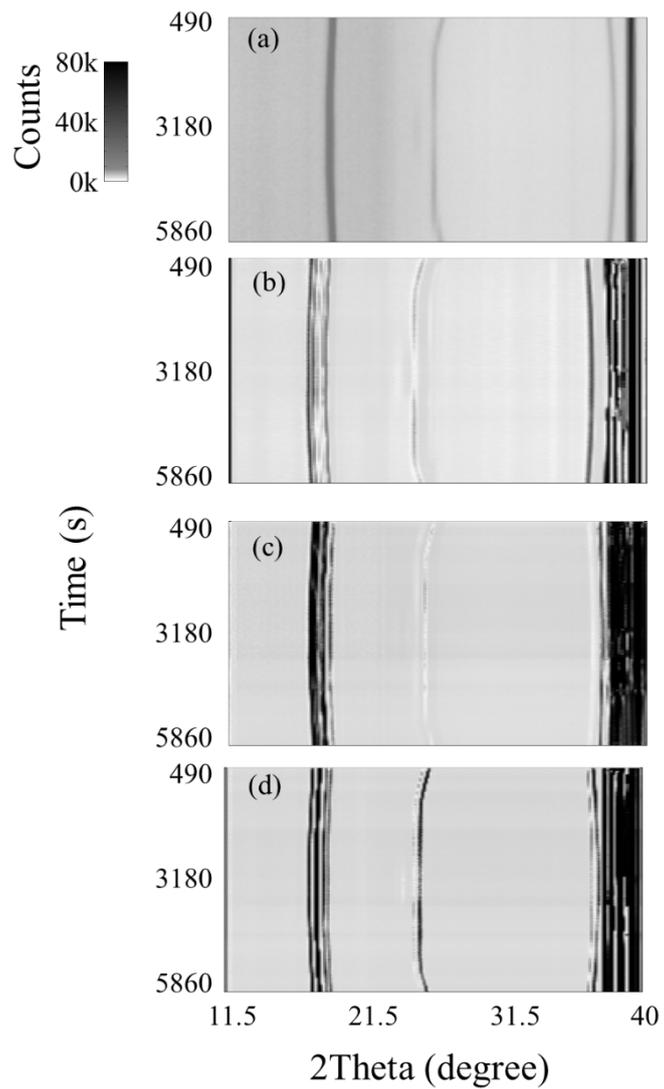

**Fig. 4.** XRD patterns collected at 1.0 C superposed with the Visualized Attention Weight: (a) XRD pattern, and calculated results in (b) Case 1, (c) Case 2, and (d) Case 3. The black and white areas represent the maximum and minimum intensities in the XRD patterns, respectively.

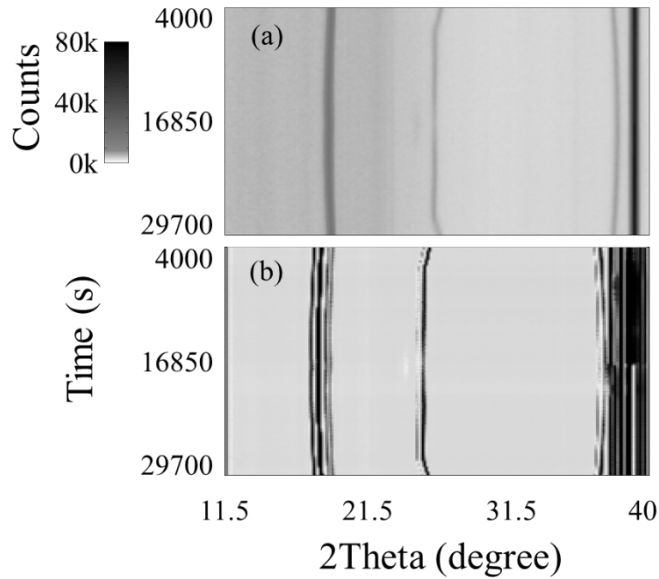

**Fig. 5.** XRD pattern collected at 0.2 C superposed with the Visualized Attention Weight: (a) XRD pattern, and (b) calculated results in Case 3. The black and white areas represent the maximum and minimum intensities in the XRD patterns, respectively.

4. **Summary**

The important peaks corresponding to the properties of lithium-ion secondary batteries were automatically extracted from in-situ X-ray diffractions obtained during a charge–discharge test of the batteries. The extraction process combined two machine learning techniques (CNN and Attention). The deep features selected the lattice constant of the cathodic active material as a predictor of the cell voltage, and the crystallographic hysteresis of the two active materials as a predictor of the charge–discharge rate. These predictions of the machine learning accord with electrochemical knowledge. By visualizing the deep features, the machine learning technique clarified the domain knowledge of materials and physical science.

As the Attention mechanism is applicable to multi-task trained models, the correlations between the spectral features and the interesting physical properties can be selectively visualized by appropriately setting the objective variables. Moreover, because the deep features are automatically extracted, this method can extract information from various measurement spectra, such as nuclear magnetic resonance, Fourier transform infrared, the Mössbauer effect, and neutron experiments.

5. **Acknowledgements**

We thank Dr. T. Tsubota, Mr. Y. Hayashi, Mr. N. Kanayama, Mr. R. Wada and Mr. K.

Kohno (KOBELCO research institute, INC.) for helpful suggestions and the experimental supports of the in-situ XRD measurements and the fabrication of the lithium ion secondary batteries.

https://www.kobelcokaken.co.jp/en/index_e.html